\documentclass[conference]{IEEEtran}
\IEEEoverridecommandlockouts

\usepackage{cite}
\usepackage{amsmath,amssymb,amsfonts}
\usepackage{algorithmic}
\usepackage{graphicx}
\usepackage{textcomp}
\usepackage{xcolor}
\def\BibTeX{{\rm B\kern-.05em{\sc i\kern-.025em b}\kern-.08em
    T\kern-.1667em\lower.7ex\hbox{E}\kern-.125emX}}

\usepackage{hyperref}
\definecolor{linkpink}{RGB}{220, 20, 100}
\hypersetup{
    colorlinks=true,
    urlcolor=linkpink,
    linkcolor=black,
    citecolor=black
}

\usepackage{gensymb}
\usepackage{amssymb}
\usepackage[utf8]{inputenc}
\usepackage{CJKutf8}
\usepackage{booktabs}
\usepackage{tabularx}  
\usepackage{multirow}  
\usepackage{rotating}
\usepackage{xcolor}
\usepackage{xspace}
\usepackage{makecell}
\usepackage{array}
\usepackage{url}
\makeatletter
\DeclareRobustCommand\onedot{\futurelet\@let@token\@onedot}
\def\@onedot{\ifx\@let@token.\else.\null\fi\xspace}
\def\eg{\emph{e.g}\onedot} 
\def\ie{\emph{i.e}\onedot} 
 
 \def\vs{\emph{vs}\onedot}
\def\wrt{w.r.t\onedot} 
\def\etal{\emph{et al}\onedot}

\IEEEoverridecommandlockouts

\usepackage{multirow}
\usepackage{tabularx}  
\usepackage[table]{xcolor}
\usepackage{makecell}

\usepackage{booktabs}

\usepackage{xspace}
\makeatletter

\definecolor{bestcolor}{RGB}{255, 230, 235}   
\definecolor{secondcolor}{RGB}{230, 240, 255} 

\definecolor{headgray}{RGB}{238,238,238}
\definecolor{oursblue}{RGB}{232,242,255}
\definecolor{oursdarkblue}{RGB}{20,78,140}

\newcommand{\ours}[1]{\textcolor{oursdarkblue}{\textbf{#1}}}

\begin{document}

\title{OcTreeMamba: Topology-Aware Mamba for 3D Medical Segmentation via Octree Sparsification}

\author{
\IEEEauthorblockN{
Kaixiang Mai\textsuperscript{1},
Jiaqi Yu\textsuperscript{1},
Yuzhu Ji\textsuperscript{1,*},
Yiqun Zhang\textsuperscript{1},
Zhijie Wang\textsuperscript{2},
Haijun Zhang\textsuperscript{3},
Yiu-ming Cheung\textsuperscript{4}
}

\IEEEauthorblockA{
\textsuperscript{1}\textit{Guangdong University of Technology, Guangzhou, China},
\textsuperscript{2}\textit{RIKEN Center for Advanced Intelligence Project, Tokyo, Japan}, \\
\textsuperscript{3}\textit{Harbin Institute of Technology, Shenzhen, China}, 
\textsuperscript{4}\textit{Hong Kong Baptist University, Hong Kong, China}
}

}

\maketitle

\begingroup
\renewcommand{\thefootnote}{}
\footnotetext{
\textsuperscript{*} Corresponding author: Yuzhu Ji

This work was supported in part by the National Natural Science Foundation of China (NSFC) under Grants 62302104 and 62476063, the Guangdong Basic and Applied Basic Research Foundation under Grants 2026A1515011337 and 2025A1515011293, and the Science and Technology Planning Project of Guangdong under Grant No. 2021B0101220006.
}
\addtocounter{footnote}{-1}
\endgroup

\maketitle

\begin{abstract}
Accurate segmentation of 3D anatomical structures is crucial to medical image analysis. Recently, Mamba-based approaches have achieved promising segmentation performance in capturing long-range dependencies with linear-time computational complexity. However, heuristic selective scanning strategies inherently disrupt the intrinsic spatial hierarchy and increase computational redundancy, ignoring the inherent sparsity of 3D anatomical structures. To this end, we propose OcTreeMamba, a hierarchical framework that incorporates Octree structures into SSM, achieving 3D topology-aware long-range context feature propagation and integration. 
Specifically, a Structure Learner performs anatomy-adaptive structured sparsification through hierarchical gating and Gumbel-Softmax sampling, enabling differentiable selection of anatomical regions. The selected regions are subsequently organized by a Dynamic Octree Builder into a multi-level topology that preserves their hierarchical connectivity. Based on the constructed Octree, a bidirectional Tree-SSM propagates and integrates contextual features along connectivity-aware leaf--parent--root and root--parent--leaf trajectories. Extensive experiments on four benchmark datasets show that our proposed OcTreeMamba achieves competitive performance \wrt topological continuity and computational efficiency, particularly for recovering the complex structure on vascular datasets. 
The code is available \href{https://github.com/MikeMak123/OcTreeMamba}{\textcolor{linkpink}{here}}.
\end{abstract}

\begin{IEEEkeywords}
Octree structure, State space model, Mamba, Dynamic scan path, Medical image segmentation
\end{IEEEkeywords}

\section{Introduction}\label{sec:intro}

Automatic segmentation of anatomical structures, such as blood vessels, airways, tumors, and organs, is a foundation task in computer-aided medical image analysis and facilitates disease diagnosis, treatment planning and surgical navigation~\cite{litjens2017survey, esteva2019guide, antonelli2022medical}. Deep Convolutional Neural Networks (DCNNs), particularly the U-Net~\cite{unet} and its variants~\cite{attention-unet,3d_unet,nnU-Net,3d_ux-net, unet++,unet3+}, have achieved remarkable progress in medical image segmentation. However, the locality of convolutional operations inherently limits their ability to capture long-range dependencies within confined receptive fields. Therefore, CNN-based models with residual connections~\cite{he2016deep}, deformable convolutions~\cite{dai2017deformable}, large kernels~\cite{peng2017large} and attention mechanisms~\cite{wang2018non} have been proposed for capturing long-range contextual features. However, it remains challenging for CNN models to segment anatomical structures, such as vessels and tumors, with intricate branching patterns, irregular shapes or elongated morphology. In contrast, Vision Transformer (ViT)-based methods~\cite{vision-transformer, swin_transformer,transunet,swinunet, SwinUNETR,timLinCXZLZ22} were proposed to model global contextual relations by using a dense query-based self-attention block. However, the quadratic computational and memory complexity in terms of the number of tokens becomes the bottleneck when applying it to 3D voxel grids. In addition, ViT-based models practically rely on sliding-window strategies and aggressive downsampling, which, to some extent, impedes the recovery of fine-grained structures with spatial details.

\begin{figure}[!t]
    \centering
    \includegraphics[width=0.9\linewidth]{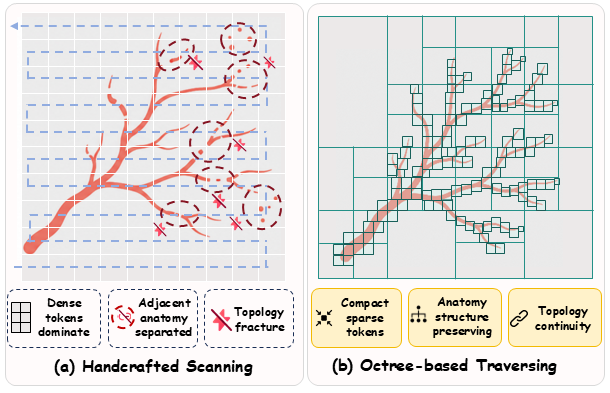}
    \caption{\textbf{Motivation of OcTreeMamba. (a) Dense scanning over 3D grids disrupts local anatomical adjacency and increases computation redundancy on non-informative background regions. (b) Anatomy-aware sparse octree topology constructed in OctreeMamba aims to propagate features along foreground structures through hierarchical interactions, improving topology preservation and structure-aware segmentation while maintaining a memory-performance balance.} }
    \label{fig:teaser}
\end{figure}

Recently, State Space Models (SSMs)~\cite{ssm1}, particularly the Mamba architecture~\cite{Mamba}, have been proposed for vision and natural language processing tasks. Due to efficiency for long-term feature propagation and linear scalability with the sequence length, SSM-based models have drawn significant attention~\cite{videomamba, s4nd}. For medical image segmentation, Mamba-based frameworks, such as U-Mamba~\cite{U-Mamba}, Vim~\cite{zhu2024vision} and VMamba~\cite{vmamba}, have been developed. To perform sequence modeling for the 3D volumetric data, these approaches practically flatten the dense spatial grids into 1D sequences. Generally, handcrafted scanning patterns are introduced to obtain rich contextual relations with dense spatial traversal. However, the handcrafted scanning with fixed patterns practically corrupts the intrinsic spatial relations and makes it fail to capture the geometry and spatial distribution of anatomical structures effectively~\cite {U-Mamba}. Moreover, adjacent voxels in the 3D non-causal volume data are widely separated when scanning the 1D sequence, causing inaccurate segmentation results with topological inconsistencies. In addition, the scanning algorithm generally propagates hidden states among voxels, without identifying whether the voxels contain valid contextual features. It thus increases a large amount of redundant computation and latency on multiple traversals for propagating features across structurally irrelevant regions, such as backgrounds and imaging noise. Furthermore, the sequential context feature propagation on dense grids leads to misalignment with the topology and sparse distribution of anatomical structures (see Fig.~\ref{fig:teaser}(a)).

We argue that, to achieve topology-preserving and computationally efficient sequence modeling, the scanning trajectories should: 1) \textit{align with the topologies of anatomical structures intrinsically}, and 2) \textit{be adaptively sparse by filtering out non-informative nodes}. Therefore, instead of introducing multiple handcrafted dense traversal trajectories, we reformulate feature propagation in SSM over a sparse tree-based topology to conform the propagation paths to the intrinsic anatomical structures (see Fig.~\ref{fig:teaser}(b)). To this end, we propose \textbf{OcTreeMamba}, a novel topology-adaptive state space model for 3D medical image segmentation. Our key technical insight is to \textit{reformulate feature propagation on a sparse yet efficient Octree topology on dense grids}. Specifically, the proposed model consists of three key modules: 1) A Structure Learner is designed to perform differentiable anatomy-adaptive sparsification through hierarchical gating and Gumbel-Softmax sampling, enabling informative anatomical regions to be selected in an end-to-end manner; 2) A Dynamic Octree Builder is proposed to organize the selected regions into a multi-level Octree topology while preserving their hierarchical connectivity; 3) A bidirectional Tree-SSM is developed to facilitate efficient feature aggregation and broadcasting along connectivity-aware bidirectional tree traversal. Experimental results on four benchmark datasets demonstrate that OcTreeMamba achieves competitive segmentation accuracy and state-of-the-art topology preservation on vascular datasets, while maintaining a favorable memory-performance balance on dense organ and lesion segmentation tasks.
In summary, our main contributions are threefold:

1) \textbf{We propose a novel Octree-based Mamba framework for 3D medical image segmentation.} To the best of our knowledge, we are the first to integrate the hierarchical Octree structure to perform feature propagation along tree topology in SSM-based models for medical image segmentation. 

2) \textbf{We develop a hierarchical Octree construction mechanism for preserving the differentiability of the framework,} while dynamically sparsifying the Octree to enable semantic-driven structure learning. 

3) \textbf{We present a topology-aware bidirectional feature propagation by adapting the Tree-SSM.} It reformulates conventional scan paths as hierarchical propagation along Octree connectivity, enabling long-range dependency modeling while preserving local topology.

\section{Related work}\label{sec:relatedwork}
\subsection{CNN-based Models}
CNN-based models, \eg, U-Net and nnU-Net~\cite{unet, nnU-Net}, have prevailed in the field of medical image segmentation due to their ability to extract representative local features, becoming foundational architectures for various medical image analysis tasks~\cite{attention-unet,unet++,kamnitsas2017efficient,bernard2018deep}. Generally, these architectures introduce skip connections~\cite{unet} and multi-scale feature fusion~\cite{unet++,unet3+}, to integrate local and global contextual features and recover spatial details from downsampled features. To process volumetric data, 3D U-Net~\cite{3d_unet} and V-Net~\cite{vnet} adapted 3D convolution operations to exploit inter-slice correlations in volumetric data. Subsequent works further augmented this paradigm by incorporating residual learning~\cite{res-unet} and attention mechanisms~\cite{attention-unet, liu2025multidirectional}, enabling the network to focus on discriminative regions while suppressing background noise. However, this efficiency stems from their reliance on local convolutions, which predict voxel-wise labels based on limited receptive fields, and thus struggle to capture long-range dependencies and global semantic context.

\subsection{Transformer-based Models}
To overcome the locality limitations of CNNs, Dosovitskiy~\etal~\cite{vision-transformer} introduced Vision Transformers (ViT) by using self-attention to model global dependencies, inspiring CNN-ViT hybrid architectures for medical image segmentation. Specifically, Chen~\etal~\cite{transunet} pioneered TransUNet\cite{transunet} by integrating a Transformer bottleneck into a CNN encoder-decoder architecture for global context modeling. Subsequently, Hatamizadeh~\etal~\cite{unetr} presented UNETR, a pure Transformer framework that processes 3D volumetric patches with a ViT encoder to capture long-range dependencies. Furthermore, Hatamizadeh~\etal~\cite{SwinUNETR} developed Swin-UNETR, incorporating shifted-window attention to reduce the cost of dense self-attention. However, applying Transformers to 3D images remains constrained by quadratic computational complexity~\cite{self-attention, unetr, vision-transformer}. Different from these approaches, OcTreeMamba avoids dense pairwise interactions over volumetric tokens by organizing informative anatomical regions into a hierarchical octree and propagating long-range structural information through the SSM, while retaining a dense convolutional pathway to preserve local details.

\begin{figure*}[!t]
    \centering
    \includegraphics[width=0.90\textwidth]{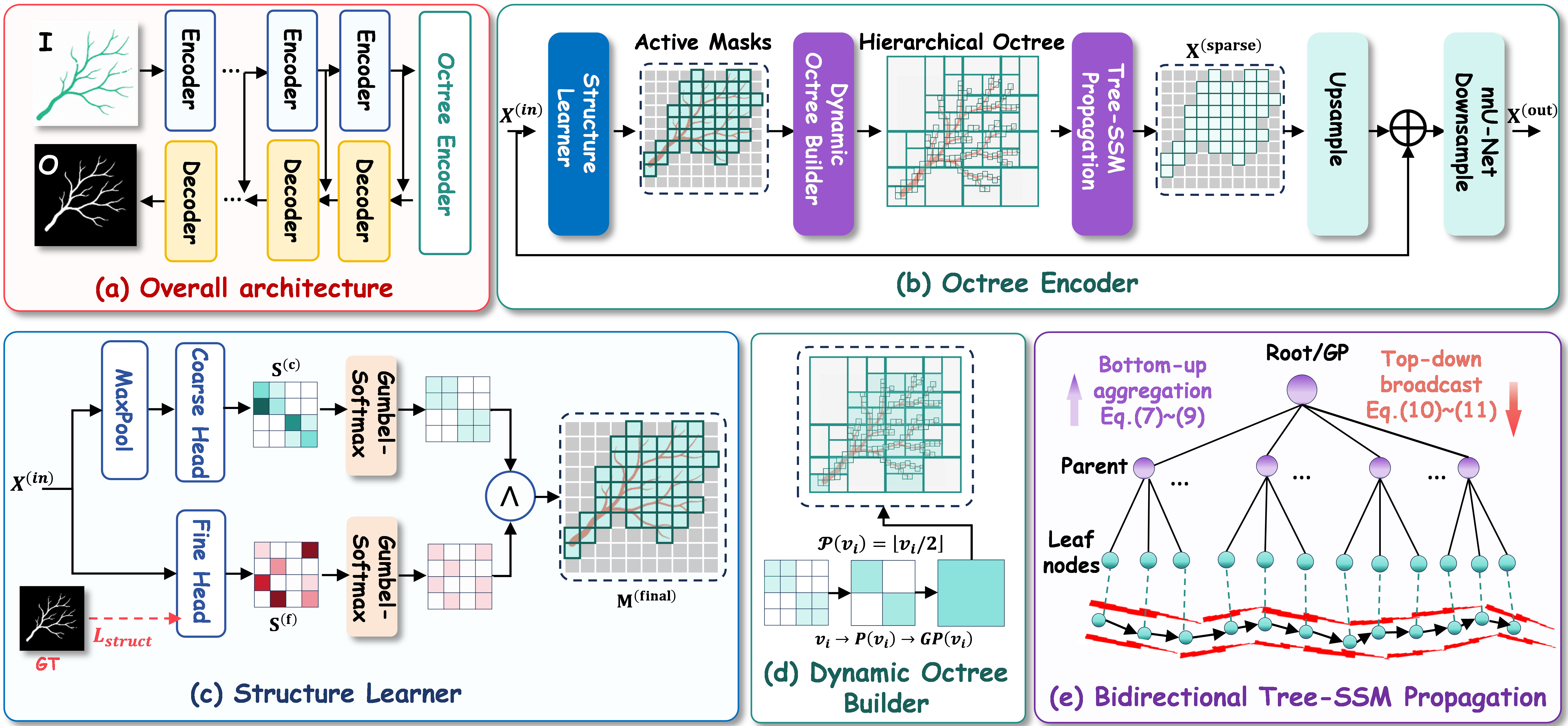}
    \caption{\textbf{Overview of the proposed \texttt{OcTreeMamba}.} The overall model (a) is built upon a UNet encoder-decoder framework. The Octree Encoder (b) is the core component, which consists of three modules: a topology-aware Structure Learner (c) for identifying active anatomical regions, a Dynamic Octree Builder (d) for constructing hierarchical sparse topology, and a bidirectional Tree-SSM Propagation module (e) for topology-aware long-range feature modeling.}
    \label{fig:overview}
\end{figure*}

\subsection{Mamba-based Models}
To capture long-range dependencies while preserving linear computational efficiency, the State Space Model (SSM) and its variants~\cite{ssm1,ssm2,ssm3} have been introduced for both vision and natural language processing tasks. Specifically, S4~\cite{ssm1} was proposed by improving the original SSM using structural constraints on the state matrix and initializing it with High-Order Polynomial Projection Operator (HIPPO)~\cite{gu2020hippo}. Gu~\etal~\cite{Mamba} developed Mamba, also known as S6, by introducing the selective scan mechanism and achieving $O(N)$ inference complexity and parallel training capability. Furthermore, considering the non-causal nature of images, Vim~\cite{zhu2024vision} and VMamba~\cite{vmamba} incorporate strategies to traverse image patches via multiple heuristic scanning paths. Recently, hybrid architectures integrating Mamba with U-Net or Transformers, such as U-Mamba~\cite{U-Mamba} and SegMamba-V2~\cite{xing2025segmamba}, have been proposed for 3D medical image segmentation by adapting the handcrafted scanning patterns to the 3D space along tri-plane perspectives. 
However, these methods still rely on predefined dense scanning paths without learning anatomy-adaptive hierarchical representations, separating adjacent 3D voxels and causing topological inconsistencies~\cite{localmamba, vmamba, grootvl}. In contrast, OcTreeMamba selects informative anatomical regions, organizes them into a hierarchical Octree, and propagates features along topology-aware traversal paths with less redundant computation over background regions.

\section{Method}\label{sec:method}
\subsection{Overview}

As shown in Fig.~\ref{fig:overview}, OcTreeMamba is built upon an encoder-decoder architecture. The core of the framework is the \textbf{Octree Encoder}, which consists of a topology-aware structure learner, a dynamic octree builder and a bidirectional Tree-SSM propagating module (Fig.~\ref{fig:overview}(c)-(e)). Specifically, the topology-aware \textbf{Structure Learner} predicts active anatomical masks from the dense feature map, and the \textbf{Dynamic Octree Builder} constructs a hierarchical octree topology leveraging the predicted semantic activation masks. The \textbf{Bi-directional Tree-SSM Propagation} module performs topology-aware feature propagation over the constructed octree to model long-range semantic dependencies among sparse anatomical regions. The obtained features are upsampled and re-injected into the dense backbone through residual fusion, followed by downsampling operations to produce the output feature map at multiple scales. This design preserves fine-grained spatial details while integrating local and global structural context through topology-aware Tree-SSM propagation.

\subsection{Structure Learner}
For 3D medical image data, anatomical regions for segmentation generally occupy only a limited portion of the volumetric space. In contrast, background regions and homogeneous interiors contribute little to long-range structural modeling. Directly applying state-space propagation to the dense grid not only introduces redundant computation but also weakens the structural responses of sparse foreground regions in sequential modeling. Therefore, before constructing the octree topology, we designed a structure learner to locate anatomy-relevant regions for subsequent topology-aware propagation.

Given an input feature $\mathbf{X}^{(in)}$, the structure learner produces activation scores at two complementary granularities, \ie, coarse- and fine-grained predictions. Specifically, the coarse predictor $\Phi^{(c)}$ produces binary activation scores $\mathbf{S}^{(c)}$:
\begin{equation}
\mathbf{S}^{(c)} = \Phi^{(c)}(\texttt{MaxPool}(\mathbf{X}^{(in)})),
\end{equation}
where $\mathbf{S}^{(c)}$ denotes the prediction scores for the inactive and active states at each spatial location. \texttt{MaxPool($\cdot$)} represents a max-pooling operation to reduce the feature resolution. This coarse prediction aims to provide a stable semantic prior for suppressing large irrelevant regions. In parallel, a fine predictor $\Phi^{(f)}$ operates on the original resolution and produces fine-grained activation map $\mathbf{S}^{(f)}$ to capture boundary details:
\begin{equation}
\mathbf{S}^{(f)} = \Phi^{(f)}(\mathbf{X}^{(in)}).
\end{equation}

To enable end-to-end optimization of discrete structure selection, a Gumbel-Softmax relaxation is introduced to provide differentiable approximations to the binary activation decisions. For each prediction granularity $r\in \{{c},{f}\}$, the relaxed activation probability is computed as:
\begin{equation}
\mathbf{P}^{(r)}_k =
\frac{\exp((\mathbf{S}^{(r)}_k+g^{(r)}k)/\tau)}
{\sum_{j}\exp((\mathbf{S}^{(r)}_j+g^{(r)}_j)/\tau)},
\quad k \in \{0, 1\}
\end{equation}

where $\mathbf{S}^{(r)}_k$ and $\mathbf{P}^{(r)}_k$ denote the activation score and relaxed probability for decision $k$ at granularity $r$. The indices $k=0$ and $k=1$ represent the inactive and active decisions. $g^{(r)}_k\sim\mathrm{Gumbel}(0,1)$ denotes Gumbel noise, and $\tau$ is the temperature coefficient. By applying the Gumbel-Softmax relaxation, two binary activation masks can be derived by selecting the most probable activation state:
\begin{equation}
\mathbf{M}^{(r)} =
\mathbb{I}\left[
\arg\max_{k\in{0,1}}\mathbf{P}^{(r)}_k=1
\right],
\quad r\in \{c,f\},
\end{equation}
where $\mathbb{I}[\cdot]$ is the indicator function. $\mathbf{M}^{(c)}$ and $\mathbf{M}^{(f)}$ denote the coarse- and fine-grained binary masks, respectively. The hard binary masks are used in the forward pass, while gradients are propagated through the corresponding soft probabilities, enabling end-to-end optimization of the discrete structure.

In practice, these two binary activation masks describe structural activation \wrt different granularities. Intuitively, the coarse mask provides region-level semantic reliability, whereas the fine mask preserves local boundary and thin-structure responses. To enforce hierarchical consistency between the two scales, we introduce a top-down gating mechanism:
\begin{equation}
    \mathbf{M}^{(final)} = \mathbf{M}^{(f)} \wedge (\texttt{Upsample}(\mathbf{M}^{(c)})),
\end{equation}
where $\wedge$ denotes element-wise logical intersection, which retains a voxel only when it is activated by both the fine-scale predictor and its corresponding coarse-scale parent, thereby suppressing isolated activations from local predictions. Such a multi-granularity compensation design is important for sparse and elongated anatomical structures, where both global support localization for the  Octree structure and local boundary preservation are required for achieving accurate segmentation.

\begin{table*}[!t]
  \caption{Quantitative validation on {ImageCAS}, {TopCoW2024}, {BraTS2021}, and {AbdomenCT} benchmark datasets. }
  \label{tab:benchmarking-4-datasets}
  \centering

    \resizebox{0.82\textwidth}{!}{%
  \begin{tabular}{c|llcccc|cccc}
    \toprule
    \rowcolor{headgray}
    Dataset & Method & Backbone
    & \multicolumn{4}{c|}{Segmentation Accuracy}
    & \multicolumn{4}{c}{Efficiency / Memory} \\
    \rowcolor{headgray}
    & &
    & Dice$\uparrow$
    & clDice$\uparrow$
    & HD95$\downarrow$
    & IoU$\uparrow$
    & TM$\downarrow$
    & IM$\downarrow$
    & Param.$\downarrow$
    & TP$\uparrow$ \\

    \midrule
     \multirow{9}{*}{\rotatebox[origin=c]{90}{ImageCAS}}
    & nnU-Net & CNN
    & 81.54 & 86.60 & 16.03 & 69.16 
    & 5563.40 & 2706.91 & 30.79 & \underline{42.55} \\

    & MedNeXt & CNN
    & 80.17 & 85.20 & 22.04 & 67.26 
    & 20959.74 & 4063.70 & \underline{5.55} & 15.55 \\

    & SegFormer3D & ViT
    & 79.39 & 85.72 & 16.80 & 66.08 
    & \textbf{2145.73} & \textbf{895.16} & \textbf{4.49} & \textbf{131.88} \\

    & UNETR++ & ViT
    & 80.42 & 85.80 & 13.67 & 67.55 
    & \underline{3995.28} & \underline{1441.55} & 42.64 & 38.42 \\

    & U-Mamba\_Bot & Mamba
    & \underline{82.11} & 87.17 & \underline{13.19} & \underline{70.00} 
    & 11175.03 & 3173.21 & 42.12 & 18.91 \\

    & SegMamba & Mamba
    & 81.22 & 85.63 & 17.42 & 68.70 
    & 16963.61 & 4454.63 & 67.36 & 9.55 \\

    & SegMamba-V2 & Mamba
    & 81.13 & 85.82 & 17.04 & 68.56 
    & 15080.17 & 5457.21 & 138.77 & 11.70 \\

    \rowcolor{oursblue}
    & \ours{OcTreeMamba\_Bot} & \ours{CNN+SSM}
    & \textbf{82.31} & \textbf{87.67} & \textbf{12.88} &\textbf{70.23 } 
    & 5942.21 & 2771.79 & 34.78 & 36.57 \\

    \rowcolor{oursblue}
    & \ours{OcTreeMamba\_Enc} & \ours{CNN+SSM}
    & 81.91 & \underline{87.18} & 14.32 & 69.67 
    & 15465.01 & 3030.85 & 40.57 & 25.14 \\
    
    \midrule

    \multirow{9}{*}{\rotatebox[origin=c]{90}{TopCoW2024}}
    & nnU-Net & CNN 
    & 65.57 & 78.61 & 25.13 & 56.97 
    & 5491.87 & 2630.60 & 30.80 & \underline{44.12} \\
    
    & MedNeXt & CNN 
    & 66.82 & 79.01 & 26.54 & \textbf{58.47}  
    & 20254.12 & 3917.88 & \underline{5.56} & 15.76 \\

    & SegFormer3D & ViT 
    & 48.80 & 70.97 & 46.67 & 40.75 
    & \textbf{2925.81} & \textbf{862.44} & \textbf{4.50} & \textbf{130.13} \\

    & UNETR++ & ViT 
    & 64.96 & 76.97 & 28.60 & 56.00 
    & \underline{4956.18} & \underline{1440.80} & 42.64 & 37.60 \\

    & U-Mamba\_Bot & Mamba 
    & 60.34 & 77.81 & 29.68 & 52.89 
    & 10879.82 & 3084.89 & 42.13 & 19.15 \\

    & SegMamba & Mamba 
    & 66.99 & \underline{79.51} & 23.95 & 58.36 
    & 16419.50 & 4313.02 & 67.36 & 9.82 \\

    & SegMamba-V2 & Mamba 
    & \underline{66.99} & 79.53 & 21.80 & 58.31  
    & 14640.62 & 5315.59 & 138.77 & 12.80 \\

    \rowcolor{oursblue}
    & \ours{OcTreeMamba\_Bot} & \ours{CNN+SSM}
    & 66.28 & 79.50 & \underline{21.22} & 57.62  
    & 6258.60 & 2698.48 & 34.79 & 37.58 \\

    \rowcolor{oursblue}
    & \ours{OcTreeMamba\_Enc} & \ours{CNN+SSM}
    & \textbf{67.17} & \textbf{80.82}   & \textbf{19.68}  & \underline{58.46}
    & 15311.03 & 2939.08 & 40.58 & 25.86 \\

    \midrule

    \multirow{9}{*}{\rotatebox[origin=c]{90}{BraTS2021}}
    & nnU-Net & CNN 
    & \textbf{84.52} & -- & \textbf{9.76} & \underline{77.28} 
    & 4887.71 & 2425.92 & 31.20 & \underline{52.77} \\

    & MedNeXt & CNN 
    & 84.42 & -- & \underline{10.38} & \textbf{77.30} 
    & 17968.72 & 3527.31 & \underline{5.55} & 18.16 \\

    & SegFormer3D & ViT 
    & 80.20 & -- & 12.78 & 70.99 
    & \textbf{1816.21} & \textbf{823.51} & \textbf{4.53}& \textbf{140.08} \\

    & UNETR++ & ViT 
    & 83.75 & -- & 10.99 & 76.39  
    & \underline{4103.96} & \underline{1470.90} & 42.65 & 38.81 \\

    & U-Mamba\_Bot & Mamba 
    & 84.27 & -- & 10.78 & 77.07  
    & 9706.80 & 2842.64 & 42.12 & 22.90 \\

    & SegMamba & Mamba 
    & 84.20 & -- & 11.81 & 76.95 
    & 14708.65 & 3984.07 & 67.42 & 11.50 \\

    & SegMamba-V2 & Mamba 
    & 84.10 & -- & 11.42 & 76.88  
    & 13183.95 & 4985.98 & 138.78 & 14.17 \\

    \rowcolor{oursblue}
    & \ours{OcTreeMamba\_Bot} & \ours{CNN+SSM}
    & 84.44 & -- & 10.82 & 77.17 
    & 5262.46 & 2501.82 & 35.19 & 42.46 \\

    \rowcolor{oursblue}
    & \ours{OcTreeMamba\_Enc} & \ours{CNN+SSM}
    & \underline{84.49}  & -- & 11.36  &  77.24 
    & 11586.51 & 2725.12 & 40.98 & 26.40 \\

    \midrule

    \multirow{9}{*}{\rotatebox[origin=c]{90}{AbdomenCT}}
    & nnU-Net & CNN 
    & 84.68 & -- & 30.26 & 77.20  
    & 4803.27 & 2163.55 & 30.71 & \underline{51.63} \\

    & MedNeXt & CNN 
    & 85.71 & -- & 28.73 & 78.41  
    & 14726.03 & 2848.82 & \underline{5.56} & 22.00 \\

    & SegFormer3D & ViT 
    & 80.08 & -- & 36.47 & 70.92 
    & \textbf{1962.25} & \textbf{649.71} & \textbf{4.50} & \textbf{133.79} \\

    & UNETR++ & ViT 
    & 86.05 & -- & \underline{21.00} & 78.96
    & \underline{4538.10} & \underline{1435.92} & 42.64 & 36.98 \\

    & U-Mamba\_Bot & Mamba 
    & \textbf{86.61} & -- & \textbf{18.20} & \textbf{79.87}  
    & 9234.50 & 2538.70 & 41.95 & 23.94 \\

    & SegMamba & Mamba 
    & 83.83 & -- & 31.35 & 76.20 
    & 14445.62 & 3901.28 & 67.36 & 11.06 \\

    & SegMamba-V2 & Mamba 
    & 86.07 & -- & 26.13 & 79.03
    & 12980.15 & 4903.85 & 138.77 & 14.08 \\

    \rowcolor{oursblue}
    & \ours{OcTreeMamba\_Bot} & \ours{CNN+SSM}
    & 85.47 & -- & 27.45 & 78.23  
    & 5539.05 & 2198.35 & 34.71 & 42.61 \\

    \rowcolor{oursblue}
    & \ours{OcTreeMamba\_Enc} & \ours{CNN+SSM}
    & \underline{86.07} & -- & 22.11 & \underline{78.98}   
    & 14011.68 & 5438.26 & 40.49 & 25.00 \\

    \bottomrule
  \end{tabular}
  }

\end{table*}

\subsection{Dynamic Octree Builder}
The activation mask produced by the Structure Learner identifies active voxels through $\mathbf{M}^{(final)}$, but these voxels remain isolated spatial locations without defining the hierarchical connections and interactions. Thus, our Dynamic Octree Builder further converts the sparse voxels into a hierarchical Octree topology. Specifically, for any active leaf node at spatial coordinate $v_i$ (where $\mathbf{M}^{(final)}(v_i) = 1$), the Octree hierarchy is established by recursively mapping it to its ancestors via coordinate quantization:
\begin{equation}
        \mathcal{P}(v_i) = \lfloor v_i / 2 \rfloor, \quad \mathcal{GP}(v_i) = \lfloor \mathcal{P}(v_i) / 2 \rfloor,
\end{equation}
where \(\mathcal{P}(v_i)\) and \(\mathcal{GP}(v_i)\) denote the parent and grandparent nodes of \(v_i\), respectively. More generally, the ancestor can be obtained by recursively quantizing the coordinates of its predecessor nodes. In this way, spatially adjacent active voxels are naturally grouped into shared parent nodes, and distant regions can interact through higher-level ancestors. In our implementation, the unstructured sparse mask is thereby transformed into a connected multi-level tree topology for the Tree-SSM message passing.

\subsection{Bidirectional OcTree-SSM Feature Propagation}
The core insight of our proposed method is to perform feature propagation via hierarchical sparse tree traversal along the learned octree. Inspired by~\cite{grootvl}, a bidirectional hierarchical Octree-topology feature propagation mechanism is developed to model long-range dependencies through bottom-up aggregation and top-down broadcasting. Intuitively, the bottom-up propagation aims to condense local responses into coarse structural states, allowing spatially separated active regions to interact through their shared ancestors. Top-down broadcasting is tailored to redistribute the contextual information to refine parent and leaf representations, while the input-conditioned channel-wise modulations selectively control information flow at each propagation stage. Consequently, long-range structural dependencies can be modelled over sparse active nodes maintaining the learned hierarchical topology.

Let $\mathcal{V}$, $\mathcal{U}$, and $\mathcal{G}$ denote the active leaf, parent, and grandparent node sets, respectively. For a leaf node $v\in\mathcal{V}$, $x_v$ denotes its local feature, $\vartheta(v)\in\mathcal{U}$ denotes its parent, and $\zeta(u)\in\mathcal{G}$ denotes the grandparent of a parent node $u$. We define the input-conditioned modulation used in different propagation stages. In the bottom-up direction, leaf features are first aggregated to their parent nodes:
\begin{equation}
\alpha_{v}^{\uparrow} = \rho(\texttt{MLP}(x_v)), \quad h_{v\rightarrow u} = \sum_{v:\vartheta(v)=u} \alpha_{v}^{\uparrow}\odot x_v,
\end{equation}
where $\rho(\cdot)$ is the SiLU activation function, \texttt{MLP}($\cdot$)) denotes a multi-layer perception for producing parameters. This vector is used for channel-wise selective modulations rather than dense transition matrices. The superscript $\uparrow$ indicates selective modulation applied in bottom-up aggregation. Similarly, the parent states are further aggregated to grandparent nodes:
\begin{equation}
\alpha_{u}^{\uparrow} = \rho(\texttt{MLP}(h_{v\rightarrow u})), \quad h_{u \rightarrow \eta}= \sum_{u:\zeta(u)=\eta} \alpha_{u}^{\uparrow}\odot h_{v\rightarrow u}.
\end{equation}
At the grandparent level, we apply an input-conditioned modulation to obtain the contextual state:
\begin{equation}
\beta_{\eta}^{\uparrow} = \rho(\texttt{MLP}(h_{u \rightarrow \eta})), \quad \tilde{h}_{u \rightarrow \eta}=\beta_{\eta}^{\uparrow} \odot h_{u \rightarrow \eta},
\end{equation}
where $\tilde{h}_{u \rightarrow \eta}$ denotes the modulated state stored at
grandparent node $\eta$. This bottom-up propagation condenses sparse local responses into coarser structural states.

In the top-down direction, let $z_{\eta}=\tilde{h}_{u \rightarrow \eta}[\zeta(u)]$ retrieve the contextual state of grandparent nodes. The contextual state is then broadcast back to parent nodes:
\begin{equation}
\beta_{u}^{\downarrow} = \rho(\texttt{MLP}(h_{v \rightarrow u})), \quad
\tilde{h}_{\eta \rightarrow u}=h_{v \rightarrow u}+\beta_u^{\downarrow}\odot z_{\eta},
\end{equation}
where $\beta_u^{\downarrow}$ gates the grandparent context $z_{\eta}$ injected into the parent state, yielding the enhanced representation $\tilde{h}_{\eta \rightarrow u}$. Each active leaf $v$ then retrieves its enhanced parent context as $z_u=\tilde{h}_{v \rightarrow u}[\vartheta(v)]$, where $\vartheta(v)$ denotes its parent. The sparse output feature is computed as:

\begin{equation}
[\alpha_{v}^{\downarrow}, \theta_{v}^{\downarrow}] = \rho(\texttt{MLP}(x_{v})), \hat{x}_v=\mathrm{Proj}\left(
\theta_v^{\downarrow} \odot \alpha_v^{\downarrow}\odot z_{u}]
\right),
\end{equation}
where $\alpha_v^{\downarrow}$ and $\theta_v^{\downarrow}$ are generated from $x_v$, and $\mathrm{Proj}(\cdot)$ denotes the layer-normalized linear projection. The enhanced sparse feature set is denoted as
$\mathbf{X}^{(sparse)}=\{\hat{x}_v\mid v\in\mathcal{V}\}$.

To integrate the sparse tree representations back into the dense backbone, we exploit a zero-initialized residual fusion strategy \cite{controlnet}. The enhanced sparse features are scattered back to the dense grid and fused with the original input:
\begin{equation}
    \mathbf{X}^{(out)} = \mathbf{X}^{(in)} + \mathbf{W} \cdot \texttt{Up}(\texttt{Scatter}(\mathbf{X}^{(sparse)})),
\end{equation}
where $\mathbf{W}$ denotes learnable parameters, which are initialized to 0. Such initialization ensures the module initially acts as an identity mapping. During the training process, $\mathbf{W}$ receives non-zero gradients when the sparse branch provides information that minimizes the loss, effectively acting as a ``soft gate'' to filter out early-stage noise. The fused tensor $\mathbf{X}^{(out)}$, now enriched with global topological semantics, is processed by the Dense Downsampling. Finally, the fused tensor $\mathbf{X}^{(out)}$ enriched with topological-aware context features is used to predict voxel-wise semantic segmentation results.

\begin{figure*}[t]
    \centering
    \includegraphics[width=0.87\textwidth]{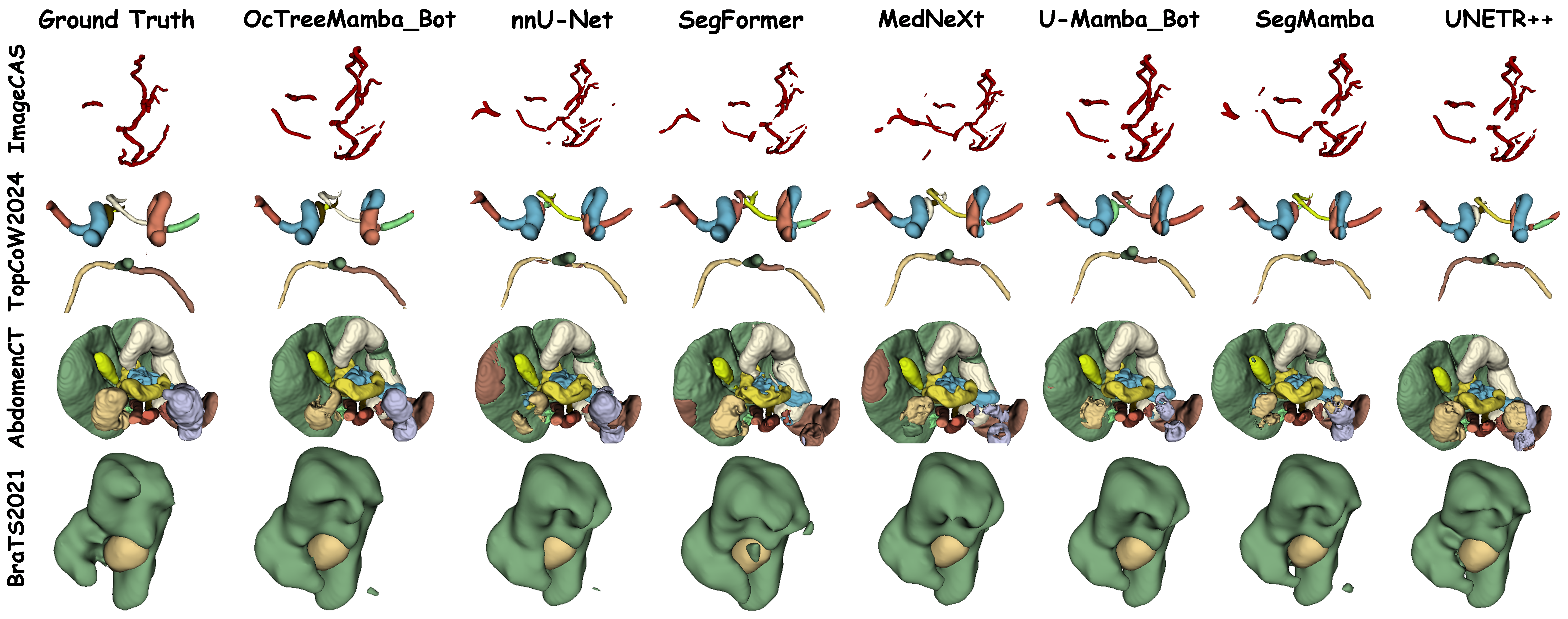}
    \caption{Quantitative comparison with state-of-the-art methods on four datasets.}
    \label{fig:visible_BIBM}
\end{figure*}

\subsection{Loss Function}\label{sec:loss}
To optimize our model, we designed a hybrid loss function that simultaneously penalizes pixel-wise segmentation errors and discrete topological structural deviations:
\begin{equation}
\mathcal{L} = \mathcal{L}_{seg} + \lambda_{struct} \mathcal{L}_{struct} + \lambda_{aux} \mathcal{L}_{aux},
\end{equation}
where $\lambda_{struct}$ and $\lambda_{aux}$ are loss weights for the structural sparsity and auxiliary deep supervision loss terms, respectively. Specifically, $\mathcal{L}_{seg}$ represents the cross-entropy loss applied to the final network output. $\mathcal{L}_{aux}$ provides auxiliary deep supervision directly to the sparse module before the outcome is fused back into the dense backbone. This auxiliary term is crucial for the Octree-SSM module, as it forces the sparse branch to rapidly capture the intrinsic topological complexities independently, without solely relying on the gradients backpropagated from the dense decoder.

\section{Experiments}\label{sec:exp}
\subsection{Datasets and Evaluation Metrics}
We performed experiments on four diverse 3D biomedical segmentation benchmarks, including ImageCAS~\cite{imagecas} for coronary artery segmentation, TopCoW24~\cite{topcow24} for circle of willis segmentation, AbdomenCT~\cite{AbdomenCT} for multi-organ abdominal segmentation, and BraTS21~\cite{brats21} for brain tumor segmentation. Segmentation performance is assessed using Dice Similarity Coefficient (DSC), Intersection over Union (IoU), and 95\% Hausdorff Distance (HD95). For the topology-sensitive vascular datasets, centerline Dice (clDice)~\cite{cldice} is additionally reported. Computational efficiency is measured by the parameter count, throughput (TP, patches/s), training and inference memory usage (TM/IM, MiB).

\subsection{Implementation Details}
We built our model upon the nnUNet framework and implemented two versions of our model, \ie, {OcTreeMamba\_Bot} and {OcTreeMamba\_Enc} by setting the octree encoder in the bottleneck only and full encoder stages, respectively. The models are optimized using {SGD} with an initial learning rate of 2e-3, a weight decay of 3e-5, and a differential polynomial learning-rate decay strategy with a power of 0.9. The learning rate of the CNN backbone is set to half that of the sparse branch. The loss weights are set to $\lambda_{\mathrm{struct}}$=0.1 and $\lambda_{\mathrm{aux}}$=0.5. We trained {OcTreeMamba\_Bot} and {OcTreeMamba\_Enc} for 500 and 1,000 epochs, respectively. All experiments are conducted on a workstation with four NVIDIA RTX 4090 GPUs.

\subsection{Main Results}
We compare our method with state-of-the-art models across three paradigms: 
(1)~\textbf{CNNs} (including {nnU-Net}~\cite{nnU-Net}, {MedNeXt}~\cite{mednext}); (2) \textbf{Vision Transformers} ({UNETR++} \cite{unetr++}, {SegFormer3D} \cite{segformer3d}); and (3) \textbf{Dense SSMs} ({U-Mamba} \cite{U-Mamba}). As shown in Table~\ref{tab:benchmarking-4-datasets}, OcTreeMamba demonstrates better performance on topology-sensitive vascular segmentation while maintaining competitive results on tumor and multi-organ benchmarks. On the ImageCAS dataset, {OcTreeMamba\_Bot} outperforms all compared methods, improving over nnU-Net by 0.77 Dice, 1.07 clDice, and 1.07 IoU, while reducing HD95 from 16.03 to 12.88. It indicates that bottleneck-level octree modeling is particularly effective for thin coronary artery structures without substantially increasing memory cost.
On the TopCoW2024 dataset, {OcTreeMamba\_Enc} achieves the best performance, improving Dice by 1.6 and reducing HD95 by 5.45 compared with nnU-Net. On BraTS2021 and AbdomenCT, OcTreeMamba shows competitive accuracy compared with CNN, Transformer, and Mamba baselines. In particular, {OcTreeMamba\_Bot} offers a better memory-performance balance, achieving substantially lower memory cost and higher throughput than dense Mamba-based methods while preserving comparable segmentation accuracy.

The qualitative comparisons in Fig.~\ref{fig:visible_BIBM} further visualize these observations. On ImageCAS and TopCoW2024, OcTreeMamba produces more continuous vascular branches with fewer isolated false positives and class inconsistencies. On AbdomenCT, it better preserves both large-organ boundaries and small anatomical structures, while on BraTS2021 it maintains a coherent tumor shape comparable to the strongest baselines. These results demonstrate that the proposed hierarchical sparse modeling is effective across both curvilinear vascular structures and compact volumetric targets.

\begin{table*}[!t]
  \caption{Ablation study of different settings of OcTreeMamba block on the ImageCAS dataset.}
  \label{tab:ablation-component-imagecas}
     \centering

  \resizebox{0.80\textwidth}{!}{%
  \begin{tabular}{llcccc|cccc}
    \toprule
    \rowcolor{headgray}
    \cellcolor{headgray}Component
    & \cellcolor{headgray}Configuration
    & \multicolumn{4}{c|}{\cellcolor{headgray}Segmentation Accuracy}
    & \multicolumn{4}{c}{\cellcolor{headgray}Efficiency / Complexity} \\

    \rowcolor{headgray}
    \cellcolor{headgray}
    & \cellcolor{headgray}
    & \cellcolor{headgray}Dice$\uparrow$
    & \cellcolor{headgray}clDice$\uparrow$
    & \cellcolor{headgray}HD95$\downarrow$
    & \cellcolor{headgray}IoU$\uparrow$
    & \cellcolor{headgray}TM$\downarrow$
    & \cellcolor{headgray}IM$\downarrow$
    & \cellcolor{headgray}Param.$\downarrow$
    & \cellcolor{headgray}TP$\uparrow$ \\
    \midrule

    nnU-Net & --  
    & 81.54 & 86.60 & 16.03 & 69.16 
    & 5563.40 & 2706.91 & 30.79 & 42.55 \\

    \midrule

    Octree + MLP & Bottleneck  
    & 81.99 & 87.36 & 12.29 & 69.78  
    & 5942.70 & 2775.52 & 34.95 & 36.66 \\

    Octree + Conv & Bottleneck  
    & 81.93 & 87.28 & 12.52 & 69.71  
    & 5932.17 & 2763.18 & 34.03 & 37.07 \\

    Octree + Self-Attn & Bottleneck  
    & 82.20 & 87.66 & 11.70 & 70.08  
    & 5937.81 & 2768.66 & 34.44 & 36.40 \\

    \rowcolor{oursblue}
    \ours{Octree + Tree-SSM} & \ours{Bottleneck}
    & 82.31 & 87.67 & 12.88 & 70.23 
    & 5942.21 & 2771.79 & 34.78 & 36.57 \\

    \midrule

    Octree + MLP & Encoder   
    & 81.63 & 87.18 & 11.87 & 69.24  
    & 11028.60 & 3034.45 & 41.05 & 26.51 \\

    Octree + Conv & Encoder  
    & 73.24 & 79.09 & 18.54 & 58.32   
    & 9227.00 & 3005.21 & 38.41 & 26.33 \\

    Octree + Self-Attn & Encoder  
    & 82.13 & 87.40 & 12.73 & 69.98  
    & 12075.42 & 3027.99 & 39.58 & 24.37 \\

    \rowcolor{oursblue}
    \ours{Octree + Tree-SSM} & \ours{Encoder}
    & 81.77 & 87.14 & 13.59 & 69.48  
    & 14957.68 & 3042.41 & 40.57 & 24.80 \\

    \bottomrule
  \end{tabular}%
  }

\end{table*}

\subsection{Ablation Study}\label{sec:abla}
\begin{figure}[!t]
    \centering
    \includegraphics[width=0.93\linewidth]{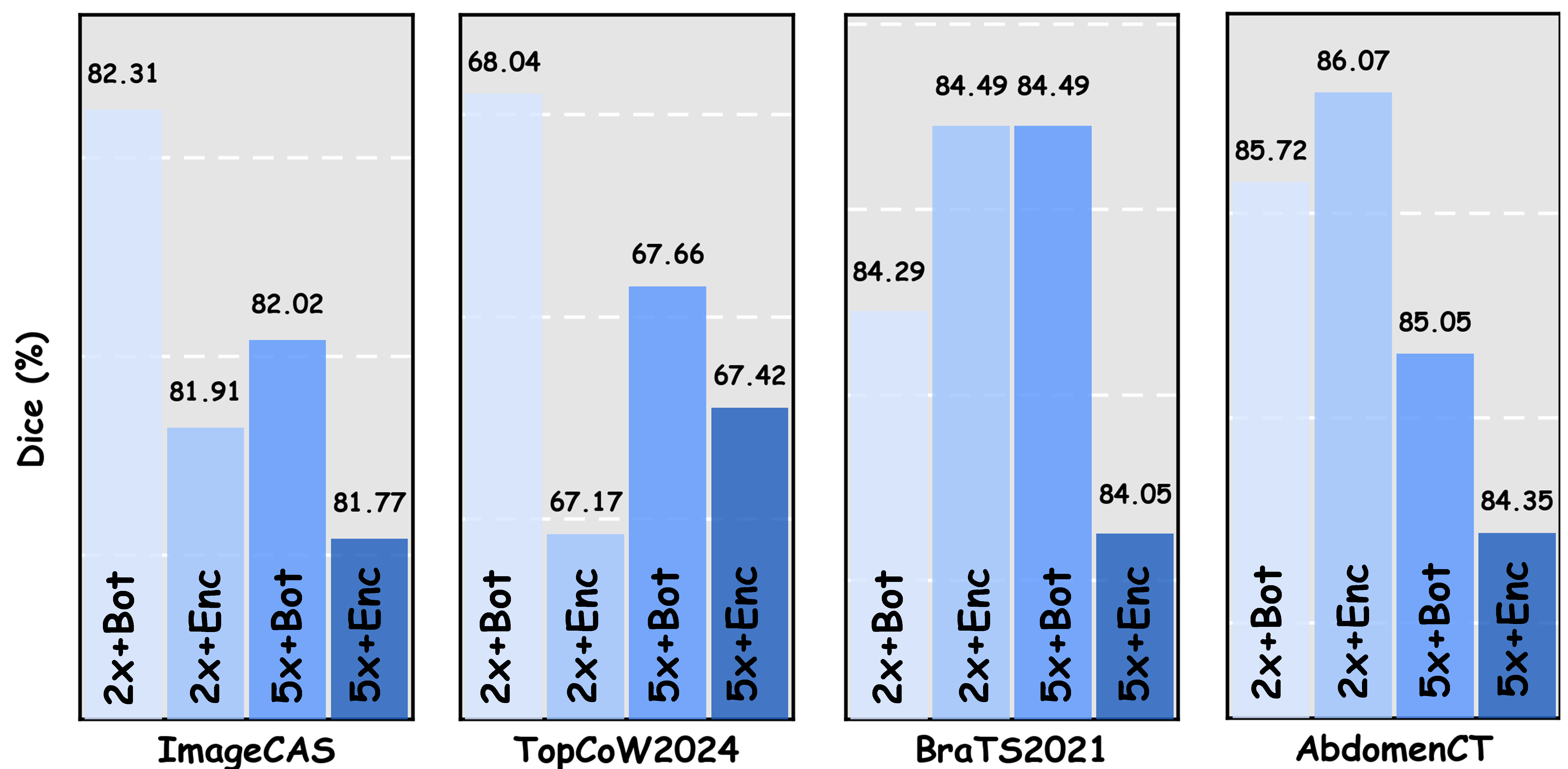}
    \caption{Ablation on different configurations of OcTreeMamba modules and learning rate strategy.}
    \label{fig:bar}
\end{figure}

We conducted ablation studies on the ImageCAS dataset to further analyze the effectiveness of OcTreeMamba, including the sparse traversing operator and different configurations of the Octree Encoder. Table~\ref{tab:ablation-component-imagecas} compares different operators within the octree-based traversing framework under both bottleneck-only and encoder-wide configurations, while Fig.~\ref{fig:bar} further evaluates the interaction between placement and learning rate scaling across different datasets. These studies aim to identify an effective accuracy-efficiency trade-off and examine how octree-based sparse modeling behaves under different architectural and optimization configurations.

\begin{figure}[!t]
    \centering
    \includegraphics[width=0.93\linewidth]{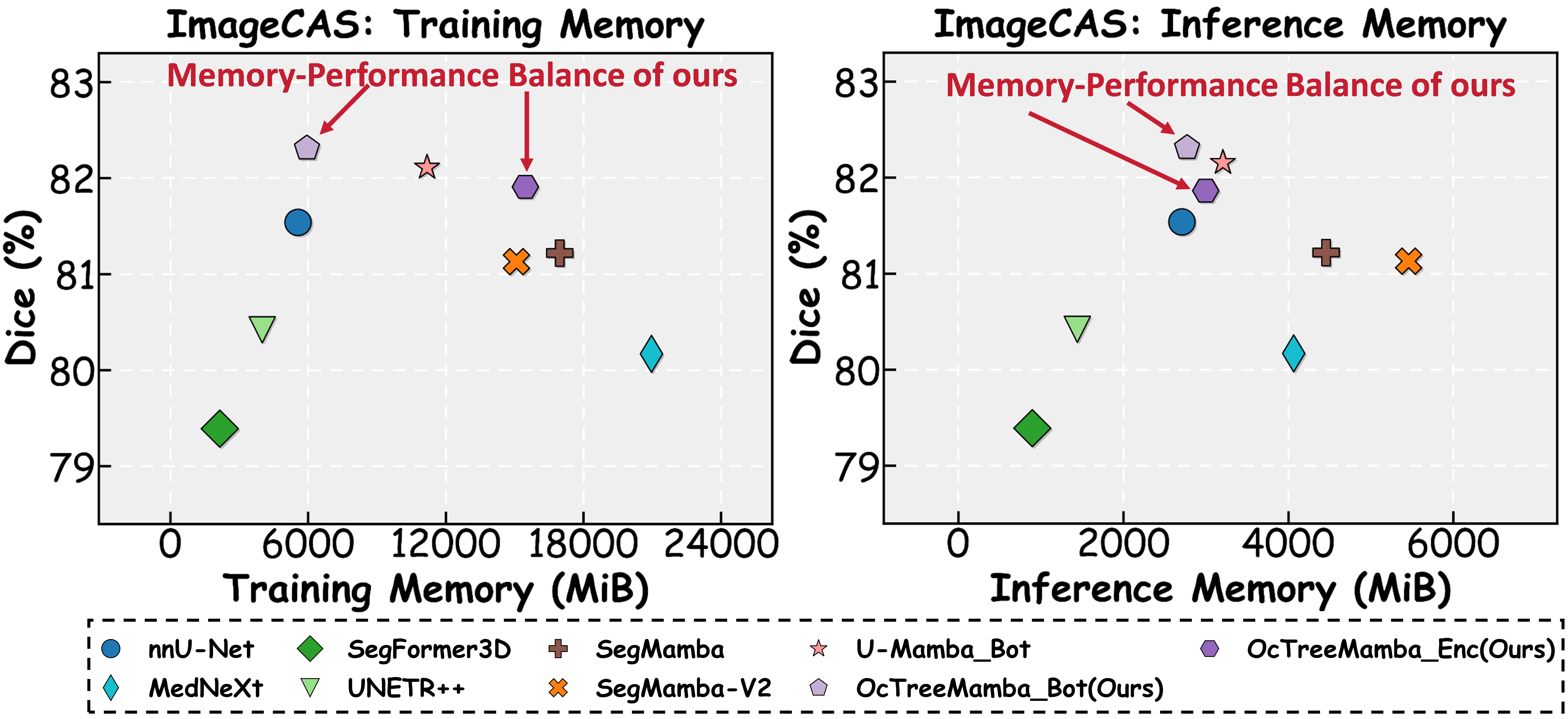}
    \caption{Memory-performance analysis on training and inference memory cost(TM, IM) and 3D segmentation performance (Dice) on ImageCAS.} 
    \label{fig:Pareto(all)}

\end{figure}

\paragraph{Ablation on the proposed OcTreeMamba block} We ablate different feature aggregation strategies of OctreeMamba. As shown in Table~\ref{tab:ablation-component-imagecas}, by replacing the feature aggregation with a point-wise MLP or convolution, the bottleneck version of baseline models improves upon the {nnU-Net} baseline, indicating that the octree-based sparse representation is more effective for modelling hierarchical anatomical structures. The performance drop of (+Conv) with 73.24 Dice may suggest convolutional operations lack an adequate receptive field for modelling long-range dependencies. In contrast, Self-Attention and our Tree-SSM, achieved superior performance. Specifically, Octree+Tree-SSM achieves superior performance on regional overlap of segmentation mask (Dice: 82.31), topological completeness (clDice: 87.67), and IoU (70.23). Although Self-Attention obtains the lowest HD95, Tree-SSM provides a more balanced performance across overlap, topology structure preservation, and efficiency-related metrics. 

\paragraph{Ablation on different module configurations and learning rate strategy} 
Fig.~\ref{fig:bar} illustrates the impact of different model configurations (\ie, {Bottleneck} \vs. {Encoder}) and the different learning rate ratio ($\gamma=2$ \vs. $\gamma=5$). For module configuration, distributing the sparse OcTreeMamba modules across the encoder ({Encoder}) yields a Dice improvement on the multi-organ {AbdomenCT} dataset (86.07 vs. 85.72), but it decreases performance on {ImageCAS}. The superiority of {OcTreeMamba\_Enc} on the {AbdomenCT} dataset indicates that multi-scale context is beneficial for capturing complex spatial relationships of large organs. Conversely, for the sparse vascular dataset, \ie, {ImageCAS}, applying sparse long-range modelling in the bottleneck layer is more effective than introducing it into high-resolution shallow layers. We believe that for curvilinear targets (\ie, {ImageCAS}, {TopCoW2024}), avoiding early-stage global interactions prevents interference from low-level background noise while maintaining higher computational efficiency. For the Learning Rate Strategy, increasing the ratio to $\gamma=5$ provides metric gains on {BraTS2021} under the {Bottleneck} setting. However, it shows performance drops on the {AbdomenCT}. Therefore, $\gamma=2$ is adopted as the default setting in our experiment. Overall, {OcTreeMamba\_Bot} and {OcTreeMamba\_Enc} are retained as two complementary configurations for different anatomical structures and hardware constraints.

\subsection{Efficiency Analysis}\label{sec:Efficiency Comparison}
As shown in Table~\ref{tab:benchmarking-4-datasets}, {OcTreeMamba\_Bot} achieves a favorable memory-performance balance, maintaining competitive or superior accuracy with fewer resources than dense Mamba-based models. Across the four datasets, {SegMamba} requires 2.6$\sim$2.9$\times$ higher training memory and 1.6$\sim$1.8$\times$ higher inference memory, while delivering 3.7$\sim$3.9$\times$ lower throughput. {SegMamba-V2} further uses about 4.0$\times$ more parameters with lower throughput. Compared with {U-Mamba\_Bot}, {OcTreeMamba\_Bot} reduces training memory by 40$\sim$47\% and provides 1.8$\sim$2.0$\times$ higher throughput. Fig.~\ref{fig:Pareto(all)} illustrates this balance. These gains come from applying sequence modeling to Octree-based sparse representations rather than dense 3D feature maps. Although {SegFormer3D} achieves the lowest memory cost and highest throughput, it suffers from substantial accuracy degradation on challenging datasets, especially TopCoW2024. In contrast, {OcTreeMamba\_Bot} preserves strong accuracy with only a small memory increase over {nnU-Net}, while {OcTreeMamba\_Enc} requires higher memory cost but provides stronger multi-scale modeling.

\section{Conclusion}\label{sec:conclusion}

In this paper, we propose {OcTreeMamba}, a novel framework for 3D medical image segmentation by introducing a hierarchical Octree structure for feature propagation along tree topology in the SSM.
Specifically, a structure learner is proposed to enable differentiable, anatomy-adaptive sparsification using Gumbel-Softmax sampling and hierarchical gating. A Dynamic Octree Builder is designed to organize the selected anatomical regions into a hierarchical Octree representation.
A bi-directional Tree-SSM is developed to facilitate Leaf-Parent-Root propagation along the OcTree-Topology-aware trajectory aligned with Octree connectivity. 
Extensive evaluation across four datasets demonstrates that {OcTreeMamba} achieves competitive segmentation accuracy and state-of-the-art performance in preserving complex vessel topology on vascular datasets and exhibits a highly competitive memory-performance balance for large-scale organ segmentation on dense anatomical targets while maintaining a low GPU memory consumption.


\bibliographystyle{IEEEtran}
\bibliography{references}

\vspace{12pt}

\end{document}